\documentclass[letterpaper, 10 pt, conference]{ieeeconf}  % Comment this line out if you need a4paper
\usepackage{cite}
\usepackage{amsmath,amssymb,amsfonts}
\usepackage{algorithmic}
\usepackage{graphicx}
\usepackage{textcomp}
\usepackage{xcolor}
\usepackage{float}
\def\BibTeX{{\rm B\kern-.05em{\sc i\kern-.025em b}\kern-.08em
    T\kern-.1667em\lower.7ex\hbox{E}\kern-.125emX}}
\usepackage[colorlinks,linkcolor=blue,anchorcolor=blue,citecolor=blue]{hyperref}
\usepackage{url}
\usepackage{footmisc}
\usepackage{times}
\usepackage{latexsym}
\usepackage{amsmath}
\usepackage{algorithm}  
\usepackage{algorithmic} 
\usepackage{graphicx}
\usepackage{multirow}
\usepackage{amssymb}
\usepackage{arydshln}
\usepackage[T1]{fontenc}
\usepackage[utf8]{inputenc}

\usepackage{microtype}

\usepackage{inconsolata}
\usepackage{CJKutf8}
\usepackage{stfloats}

\IEEEoverridecommandlockouts                              % This command is only needed if 
\title{\LARGE \bf
RPN: A Word Vector Level Data Augmentation Algorithm in Deep Learning for Language Understanding
}

\author{ Zhengqing Yuan$^{1,3}$, Xiaolong Zhang$^{1}$, \\ Yue Wang$^{1}$, Xuecong Hou$^{1}$, Huiwen Xue$^{2}$,  Zhuanzhe Zhao$^{1,4,*}$ and Yongming Liu$^{1,5,*}$
\thanks{*Corresponding Authors}% <-this % stops a space
\thanks{$^{1}$School of Artificial Intelligence, Anhui Polytechnic University, Wuhu 241000 China
        }%{\tt\small albert.author@papercept.net}
\thanks{$^{2}$School of Optoelectronic Science and Engineering, Soochow University, Suzhou 215000 China
        }%
\thanks{$^{3}${\tt\small Zhengqingyuan@ieee.org}}
\thanks{$^{4}${\tt\small Zhuanzhe727@ahpu.edu.cn}}
\thanks{$^{5}${\tt\small liuyongming1015@163.com}}
}

\begin{document}

\maketitle
\thispagestyle{empty}
\pagestyle{empty}

%%%%%%%%%%%%%%%%%%%%%%%%%%%%%%%%%%%%%%%%%%%%%%%%%%%%%%%%%%%%%%%%%%%%%%%%%%%%%%%%
\begin{abstract}

Data augmentation is a widely used technique in machine learning to improve model performance. However, existing data augmentation techniques in natural language understanding (NLU) may not fully capture the complexity of natural language variations, and they can be challenging to apply to large datasets. This paper proposes the \textbf{R}andom \textbf{P}osition \textbf{N}oise (\textbf{RPN}) algorithm, a novel data augmentation technique that operates at the word vector level. RPN modifies the word embeddings of the original text by introducing noise based on the existing values of selected word vectors, allowing for more fine-grained modifications and better capturing natural language variations. Unlike traditional data augmentation methods, RPN does not require gradients in the computational graph during virtual sample updates, making it simpler to apply to large datasets. Experimental results demonstrate that RPN consistently outperforms existing data augmentation techniques across various NLU tasks, including sentiment analysis, natural language inference, and paraphrase detection. Moreover, RPN performs well in low-resource settings and is applicable to any model featuring a word embeddings layer. The proposed RPN algorithm is a promising approach for enhancing NLU performance and addressing the challenges associated with traditional data augmentation techniques in large-scale NLU tasks. Our experimental results demonstrated that the RPN algorithm achieved \textbf{state-of-the-art} performance in all seven NLU tasks, thereby highlighting its effectiveness and potential for real-world NLU applications.

\end{abstract}

%%%%%%%%%%%%%%%%%%%%%%%%%%%%%%%%%%%%%%%%%%%%%%%%%%%%%%%%%%%%%%%%%%%%%%%%%%%%%%%%
\section{Introduction}
Data augmentation has been shown to be effective in improving the performance of various natural language processing tasks \cite{10.5555/3495724.3496249,PGD}. Often used in sentence-level tasks such as text classification \cite{10.1145/3439726} and textual inference \cite{10.5555/1619499.1619510}, data augmentation methods have consistently demonstrated improved model performance. While many methods directly modify samples, our approach aims to enhance samples by leveraging the word embeddings output from the model, as demonstrated in TinyBERT \cite{jiaoetal2020tinybert}. This allows for the utilization of various continuous probability distributions for data modifications.

There are currently two methods available for directly modifying the output data of word embeddings. The first method involves generating a perturbation tensor with elements drawn from a uniform distribution within a specific range, which is then added to the original data. This perturbed data is then used for training, and its gradient is calculated through backpropagation. For instance, the official code in FreeLB adversarial training \cite{Zhu2020FreeLB:} adopts this approach. The second method involves replacing a word in the original sentence with the closest word in the vector space, which is determined by measuring the cosine similarity between the word embeddings. This method has been used for data augmentation in various applications, such as in the work of Wang et al. \cite{wangyang2015thats} on tweet corpus. Both of these methods have been shown to effectively improve the performance of models on specific tasks.

Several issues limit the data augmentation approach described above. The first problem is that the generated perturbation is so pronounced that it directly changes all word vectors of the sentence, altering the expected value $\mathbb{E}(\mathcal{D})$ and variance of the entire sample too much. Even though the random number added to each word vector is small, it can cause a relatively significant change in the meaning of the entire sentence. Adversarial training is an example of this type of method. The second problem is that some methods lack sufficient randomness to make a strong contrast with the original sample. For instance, back-translation \cite{sennrichetal2016improving,zhang2020back}, Easy Data Augmentation (EDA) \cite{weizou2019eda}, Augmented Easy Data Augmentation (AEDA) \cite{karimietal2021aedaeasier}, and random spelling error methods directly use the original sample to make changes, which can be time-consuming when dealing with large datasets and cannot fully utilize the powerful performance of GPU or TPU.

In this paper, we propose a new data augmentation method, called the Random Position Noise algorithm (RPN), which addresses the three limitations mentioned earlier. Unlike traditional adversarial training methods, our approach does not require the gradient of the training data in the computational graph when updating the virtual samples, which improves the stability and efficiency of the method. The RPN algorithm is simple, yet more effective than traditional methods as it does not destroy the meaning of all words in the sentence nor provide insufficient randomness. The contributions of this paper are summarized as follows:

\begin{itemize}

\item We demonstrate that our algorithm achieves \textbf{state-of-the-art} performance for most sentence-level tasks, and it is compatible with a wide range of models.

\item Our method does not require frequent communication among computational nodes in large-scale training, as it does not rely on traditional adversarial training methods that require frequent gradient computation.

\item We will make the source code of all the tasks in our paper available on GitHub for readers to test and optimize our algorithm.\footnote{\url{https://github. com/DLYuanGod/RPN}}

\end{itemize} 

\section{Background}

In this section, we briefly overview a few common methods of data augmentation and the selected methods we will use in the following section. 

\subsection{Sample Augmentation}

Sample augmentation was initially widely used in computer vision to enhance task performance through techniques such as grayscale transformation, image flipping, and filtering \cite{9156446,9156559}. However, these methods may not be directly applicable to natural language processing (NLP) tasks, given that language is inherently temporal and discrete. Unlike images, language operates through sequences of discrete symbols and has grammatical and semantic structures that must be preserved. Therefore, researchers have developed specialized augmentation techniques for NLP, such as word replacement, synonym insertion, and back-translation \cite{zhang2020back,perez2014automatic,karpukhin2021denoising}.

\subsection{The EDA Method}

The EDA method is a classical approach to data augmentation, which includes Synonym Replacement, Random Insertion, Random Exchange, and Random Deletion. However, in this paper, we propose a different approach that involves random modification, as opposed to directly modifying the samples as in EDA. 

\subsection{The AEDA Method}

The AEDA method is a data augmentation approach that involves adding punctuation to the middle of a sentence. This simple yet effective technique can improve the quality of the data \cite{karimietal2021aedaeasier}.

\subsection{Back Translation}

The Back Translation approach involves using machine translation to generate new pieces of text by paraphrasing, thus expanding the dataset. One advantage of this approach is that it introduces minimal changes to the original sentence's meaning \cite{sennrichetal2016improving}. Xie et al. \cite{10.5555/3495724.3496249} applied this method to train a semi-supervised model using just over 20 data samples.

\subsection{Adversarial Training}
Adversarial training, including PGD \cite{madry2018towards}, and FGSM \cite{DBLP:journals/corr/GoodfellowSS14}, is known as the \textit{ultimate data augmentation method} and is now used extensively in natural language understanding tasks such as FreeLB adversarial training because of the consistent improvement it brings to results. 
The idea of adversarial training is to add linear perturbations in the gradient direction, as equations (\ref{FGSMeq}), tiny random perturbations are applied, see Section \ref{RPN_algorithm}. 

\begin{eqnarray}\label{FGSMeq}
\boldsymbol{\delta _{t+1}} = {\textstyle \prod_{\left \|\boldsymbol{ \delta} \right \|_{F} \le \boldsymbol{ \epsilon } }} 
\left (\boldsymbol{ \delta_{t}}+ \alpha g\left ( \boldsymbol{ \delta_{t} }\right ) / \left \|\boldsymbol{ \delta_{t}} \right \|_{F} \right ) 
\end{eqnarray}

The essence of the FreeLB method is to initialize a tensor that is uniformly distributed over a small range (e. g. -1e-4 to 1e-4)($\boldsymbol{\delta _{0}} \sim U\left ( -1e-4,1e-4 \right ) $). The generated tensor is then added to the output data of the word embeddings layer, which is then taken for normal training and backpropagation ($g\left ( \boldsymbol{ \delta_{t} }\right )$), and its corresponding gradient is computed. The next perturbation is updated as the product of the gradient and the learning rate plus the initialized value.

\section{RPN}\label{RPN_algorithm}

This section introduces the RPN method, including its theoretical implementation, pseudocode, and application examples. 

\subsection{Theoretical Implementation}
Many NLP models utilize multiple types of embeddings, such as word embeddings, segmental embeddings, and positional embeddings, to capture different aspects of the input data. In contrast, our proposed RPN method only operates on the output data of the word embeddings.

For a sentence with $n$ words, the $\boldsymbol{Z}=\left(\boldsymbol{z}_{1} , \boldsymbol{z}_{2} ,\cdots , \boldsymbol{z}_{n}\right )^{T} $ tensor denotes the one-hot representation. The original word embeddings tensor $\boldsymbol{X}_{t=0}$ is derived from Equation (\ref{eq:embeddings}). Among them, $\boldsymbol{V}=\left (\boldsymbol{v}_{1}, \boldsymbol{v}_{2},\cdots,\boldsymbol{v}_{m}\right ) $ is the learnable weight in the word embeddings layer, and $m$ denotes the word list size. 

\begin{equation}\label{eq:embeddings}
\boldsymbol{X}_{t=0} =  \left \{ x_{ij} \right \}  = \boldsymbol{Z}\boldsymbol{V}
\end{equation}

The RPN algorithm generates virtual samples for the next step ($t>0$) as shown below. We generate a random 0-1 position tensor as shown in Equation (\ref{eq:position}). This tensor is used to help us randomly select the direction of a dimension in the word vector.

\begin{equation}\label{eq:position}
\boldsymbol{P} = \begin{bmatrix}  
  1 & 0 & 1 & \cdots & 0 \\  
  0 & 1 & 0 & \cdots & 1 \\  
  \vdots & \vdots & \vdots & \ddots & \vdots \\  
  1 & 1 & 0 & \cdots & 0  
\end{bmatrix}_{n\times m}
\end{equation}

We obtain the new $\boldsymbol{P}^{\prime}$ matrices by doing the same row random transformation of the $\boldsymbol{P}$ matrices. Finally, the new sample tensor $\boldsymbol{X}_{t}$ is obtained using Equations (\ref{eq:thend}).

\begin{equation}\label{eq:thend}
\boldsymbol{X}_{t} = \boldsymbol{X}_{t-1} - \boldsymbol{P}^{\prime} \circ \boldsymbol{X}_{t-1} + \boldsymbol{X}_{t-1} \circ \boldsymbol{P}
\end{equation}

\begin{figure}[htbp]
\centering
\includegraphics[scale=0.49]{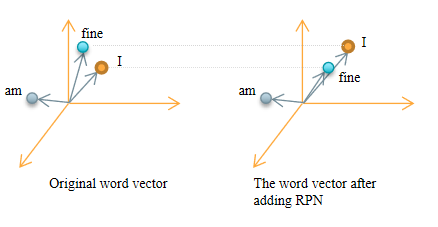}
\caption{For example, in \textit{I am fine}, the word vectors \textit{I} and \textit{fine} modifiy the values of the third dimension. }
\label{RPN_figure}
\end{figure}

As shown in Figure \ref{RPN_figure}, the purpose of adding the RPN is to replace one or more dimensions of a word vector with the value of some dimension of some other word vector. 

\begin{eqnarray}\label{RPNeq}
\min \limits_{\boldsymbol{\theta}} \mathbb{E}_{(\boldsymbol{Z}, y) \sim \mathcal{D}}\left[\frac{1}{K+1} \sum_{t=0}^{K}  L\left(f_{\boldsymbol{\theta}}\left(\boldsymbol{X}_{t}\right), y\right)\right]
\end{eqnarray}

As shown in Equation (\ref{RPNeq}), where the $f_{\boldsymbol{\theta}}\left(\boldsymbol{X}\right)$ is your language model (encoder), $\boldsymbol{\theta}$ denotes all the learnable parameters including the embeddings matrix $\boldsymbol{V}$, and $K$ is the number of times the RPN algorithm is executed.

\subsection{Pseudocode}

As shown in Algorithm \ref{RPN_al}, we input the output data $\boldsymbol{X_{0}}$ of the word embeddings, the Number of perturbations added, and the random probabilities $\varepsilon$ in 0 to 1 directly into the algorithm. 

\begin{algorithm}
	\caption{Random Position Noise (RPN)} 
	\label{RPN_al} 
	\begin{algorithmic}[1]
		\REQUIRE 
		Mini-batch $B \subset D$, word embeddings $\boldsymbol{X}_{0}$, learning rate $\tau$, number of ascent steps $K$, and probability $\varepsilon$ of modification.
		\STATE Initialize parameters $\theta$.
		\FOR{$t = 0$ to $K$}
			\STATE Initialize $g_{-1}$ to 0.
			\STATE Compute the gradient of parameters $\theta$: 
			\STATE $\quad \varrho \leftarrow \frac{1}{K+1} \mathbb{E}_{(\boldsymbol{X}, \boldsymbol{y}) \in B}\left[\bigtriangledown_{\boldsymbol{\theta}} L\left(f_{\boldsymbol{\theta}(\boldsymbol{m})}\left(\boldsymbol{X}_{t}\right), y\right)\right]$.
			\STATE Accumulate the gradient: $g_{t} \leftarrow g_{t-1} + \varrho$.
			\STATE Update the parameters: $\theta \leftarrow \theta - \tau g_{t}$.
			\STATE Initialize $\boldsymbol{\delta}$ and $\boldsymbol{P}$ to zero tensors.
			\FOR{$i = 0$ to $b \times n \times m -1 $}
				\STATE Set $\boldsymbol{P}_{\left[i \right]}$ to 1 with probability $\varepsilon$.
			\ENDFOR 
			\STATE Reshape $\boldsymbol{P}$ into a $b \times n \times m$ three-dimensional tensor. 
			\STATE Set $\boldsymbol{\delta} \leftarrow \boldsymbol{X}_{t} \circ \boldsymbol{P}$.
			\STATE Randomly shuffle the three dimensions of $\boldsymbol{\delta}^{\prime}$ and $\boldsymbol{P}^{\prime}$ equally.
			\STATE Set $\boldsymbol{X}_{t+1}^{\prime}\leftarrow \boldsymbol{X}_{t} \circ \boldsymbol{P}^{\prime}$.
			\STATE Set $\boldsymbol{X}_{t+1}\leftarrow \boldsymbol{X}_{t} - \boldsymbol{X}_{t+1}^{\prime} + \boldsymbol{\delta}^{\prime}$.
		\ENDFOR
	\end{algorithmic} 
\end{algorithm}

\subsection{Application Examples}\label{Application}
\begin{figure*}[htbp]
\centering
\includegraphics[scale=0.39]{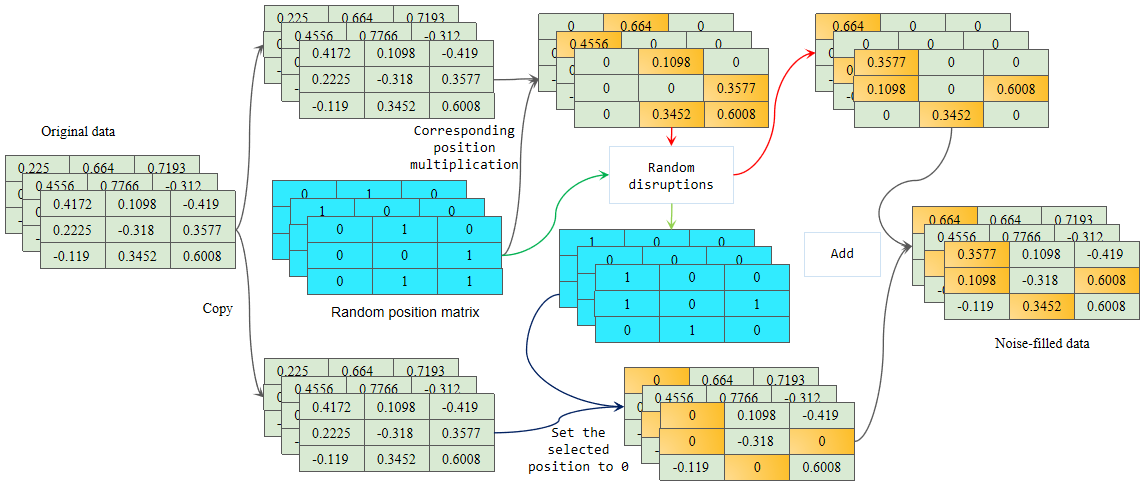}
\caption{In the real project, the original data is the word embeddings layer's output, which is a tensor in three dimensions. To illustrate, the word vectors in the figure are randomly generated. After the changes in the figure, the final Noise-filled data is the sample after adding noise. }
\label{RPN_example}
\end{figure*}

RPN can be applied to any model that includes a word embeddings layer, and is suitable for sentence-level tasks. The number of times to add noise can be specified by $K$, which theoretically increases the entropy of the data compared to the original, but may also cause the generated data to deviate further from their labels. In Figure \ref{RPN_example}, the batch size is incorporated in the actual training process, so $\boldsymbol{X}_{t}$ is a three-dimensional tensor.

The proposed approach can be adapted to various forms of adversarial training, such as FreeLB \cite{Zhu2020FreeLB:} with $N$ iterations as in Eq. (\ref{FreeLB}). Specifically, we substitute $\boldsymbol{X}+\boldsymbol{\delta}_{t}$ in the original equation with $\boldsymbol{X}_{n} + \boldsymbol{\delta}_{t}$, where $\boldsymbol{X}_{n}$ denotes the $n$-th iteration of the input during FreeLB adversarial training.

\begin{figure*}[h]
 \centering
\begin{eqnarray}\label{FreeLB}
\min _{\boldsymbol{\theta}} \mathbb{E}_{(\boldsymbol{Z}, \boldsymbol{y}) \sim \mathcal{D}, \boldsymbol{m} \sim \mathcal{M}}\left[\frac{1}{(N+1)K} \sum_{n=0}^{N} \sum_{t=0}^{K-1} \max _{\delta_{t} \in \mathcal{I}_{t}} L\left(f_{\boldsymbol{\theta}(\boldsymbol{m})}\left(\boldsymbol{X}_{n} + \boldsymbol{\delta}_{t}\right), y\right)\right]
\end{eqnarray}
\end{figure*}

\section{Experimental Setup}

In this section, we describe our experimental setup to demonstrate the performance of our algorithm. 

\subsection{Experimental Environment}
To ensure experimental fairness, the task was conducted on a homogeneous server infrastructure comprising 11 servers, each equipped with two Nvidia A5000 GPUs and utilizing a consistent batch size of 512 across all tasks.

\subsection{Experimental Model}
To demonstrate the generalizability of the RPN method, the algorithm is validated on different models. 

\subsubsection{BERT$_{BASE}$}
BERT (Bidirectional Encoder Representations from Transformers) is a large-scale neural network architecture that has achieved state-of-the-art results in various natural language processing (NLP) tasks. The BERT$_{BASE}$ model architecture includes 12 transformer blocks, 110 million parameters, and supports both masked language modeling and next sentence prediction. BERT's success has paved the way for numerous other transformer-based models in the NLP community.

\subsubsection{RoBERTa$_{BASE}$}
The RoBERTa model has emerged as one of the most remarkable models in recent years, achieving top performance on the GLUE tasks benchmark when it was first introduced \cite{wang-etal-2018-glue}. Compared to BERT, RoBERTa is trained on a richer set of pre-training data, and instead of using static masking during preprocessing, it dynamically generates a mask for each input to the model. In this experiment, we use the base version of the RoBERTa model as our baseline model, and apply the RPN algorithm for fine-tuning. The RoBERTa$_{BASE}$ model encoder \footnote{We utilize the RoBERTa-base model from the pytorch-transformers library: \url{https://github.com/huggingface/pytorch-transformers}} consists of 12 hidden layers, produces an output of 768-dimensional tensor, features 12 self-attentive heads, and has a total of 125M parameters, which are trained on English text.

\subsubsection{TextCNN}
The TextCNN model has revolutionized deep learning-based text classification and catalyzed the advancement of convolutional neural networks in natural language processing. As a fundamental deep learning model, our algorithm also contributes to this model's success.

\subsection{Dataset}
In the subsequent experiments, we chose 8 tasks, which were sentence-level tasks. 

\subsubsection{TweetEval}
TweetEval~\cite{barbieri-etal-2020-tweeteval} all formulated as multi-class tweet classification tasks. To ensure fair comparisons, all tasks are presented in a unified format with fixed training, validation, and test splits. However, due to the small size of the datasets in the sixth task, which includes six datasets, experimental results can be highly unstable. Therefore, in this study, we only focus on the first six tasks.

\subsubsection{CoLA}
The CoLA (Corpus of Linguistic Acceptability) \cite{warstadtetal2019neural}, as presented by Warstadt et al. in 2019, comprises 10,657 sentences sourced from 23 linguistic publications, and is professionally annotated by the original authors for acceptability (grammaticality). The public version of the corpus, which is provided here, includes 9,594 sentences belonging to the training and development sets, while excluding 1,063 sentences from the retention test set.

\subsubsection{SST-2}
The SST-2 (Stanford Sentiment Treebank) \cite{socheretal2013recursive} corpus, originally introduced by Socher et al. in 2013, features fully labeled parse trees that facilitate comprehensive analysis of the compositional effects of sentiment in language. Derived from the dataset presented by Pang and Lee in 2005, the corpus comprises 11,855 individual sentences extracted from movie reviews. These sentences were parsed using the Stanford parser, yielding 215,154 distinct phrases from the resulting parse trees, each of which were annotated by 3 human judges.

\section{Experimental}
In this section, we present the RPN on different tasks results. 

\subsection{RPN vs. Back-translation}
We used Baidu's translation API \footnote{\url{https://developer.baidu.com/}} to perform back-translation on the sample in order to enhance it. However, due to limitations of the API such as its batch size and word count restrictions, I was unable to conduct extensive experimentation.

\paragraph{Results} As demonstrated in Table \ref{Time Comparison}, the Baidu API has a response rate of only once every 10 seconds, resulting in a back-translation process that takes over 20 seconds to complete. However, our algorithm is able to significantly reduce this time while simultaneously improving the accuracy of on the EP tasks

\begin{table}[!h]
\centering
\caption{\label{Time Comparison}
Experimental results of five executions on the EP task with RoBERTa$_{BASE}$. 
}
\resizebox{.95\columnwidth}{!}{
\begin{tabular}{lccc}
\hline
\multirow{2}{*}{\textbf{Algorithm}} & \multicolumn{2}{c}{\textbf{EP}} & \multirow{2}{*}{\textbf{Average time(s)}} \\\cline{2-3} 
                                    & loss           & acc.            &                                        \\ \hline
Back-translation          & 1.724          & 46.94          & 21830                                \\ 
 RPN    & \textbf{1.717} & \textbf{47.11} & \textbf{1392}                       \\ \hline
\end{tabular}
}
\end{table}

\subsection{Evaluation}\label{loss}

\begin{table*}[]
\centering
\caption{\label{Reasult}
The results of the CoLA task and the six tasks in the Tweet Eval dataset. In each task, the model was trained five times to take the mean of the final results. In each task, we calculated the loss and accuracy of the model's results for five experiments on the test set. The score in the CoLA task is the score in the out-of-domain test in the Kaggle competition~\protect\footnotemark. 
}
\resizebox{.95\hsize}{!}{
\begin{tabular}{lcccccccccccccc}
\hline
\multirow{2}{*}{\textbf{Tasks}} & \multicolumn{2}{c}{\textbf{EP}} & \multicolumn{2}{c}{\textbf{ER}} & \multicolumn{2}{c}{\textbf{SA}} & \multicolumn{2}{c}{\textbf{HSD}} & \multicolumn{2}{c}{\textbf{ID}} & \multicolumn{2}{c}{\textbf{OLI}} & \multicolumn{2}{c}{\textbf{CoLA}} \\\cline{2-15}
                                & loss           & acc.            & loss           & acc.            & loss           & acc.            & loss            & acc.            & loss           & acc.            & loss            & acc.            & dev acc.         & score           \\ \hline
  \textbf{Baseline}  &&&&&&&&&&&&&&                           \\
   BERT$_{BASE}$     &  2.112          & 37.13          & 0.515          & 81.32  & 0.711          & 70.16          & 0.702  & 72.33          & 0.736          & 68.75          & 0.496           & 82.69          & 0.741           & 0.339           \\ \hdashline[0.5pt/5pt]
  \textbf{Other Methods}  &&&&&&&&&&&&&&                           \\
BERT$_{BASE}$+EDA        & 2.103          & 37.61          & 0.533          & 81.45         & 0.706          & 69.88          & 0.701  & 72.41          & 1.465          & 72.24          & 0.484           & 84.77          & 0.781           & 0.340          \\
BERT$_{BASE}$+AEDA        & 2.100          & 37.37          & 0.829          & 81.68         & 0.755          & 70.10          & 0.702  & 72.32           & 1.579          & 70.31          & 0.426           & 83.77          & 0.780           & 0.381           \\
BERT$_{BASE}$+FreeLB         & 2.102          & 37.10          & 0.733          & 81.53         & 0.663          & 70.73          & 0.684  & 72.35          & \textbf{0.726}          & 70.18          & \textbf{0.369}           & \textbf{84.77}          & 0.772           & 0.392           \\ \hdashline[0.5pt/5pt]
\textbf{Ours}  &&&&&&&&&&&&&&                           \\
BERT$_{BASE}$+RPN            & \textbf{2.067} & \textbf{37.74} & \textbf{0.551} & \textbf{81.82}          & \textbf{0.659} & \textbf{70.80} & \textbf{0.658}           & \textbf{74.01} & 1.751 & \textbf{72.27} & 0.378  & 84.74 & \textbf{0.799}  & \textbf{0.501}  \\ \hline 
  \textbf{Baseline}  &&&&&&&&&&&&&&                           \\
  RoBERTa$_{BASE}$         &  1.755          & 46.11           & 0.622          & 81.45  & 0.650          & 70.51          & 0.687  & 73.21          & 0.616           & 73.83          & 0.428           & 83.20          & 0.772           & 0.414           \\ \hdashline[0.5pt/5pt]
  \textbf{Other Methods}  &&&&&&&&&&&&&&                           \\
RoBERTa$_{BASE}$+EDA        & 1.724         & 47.05          & 0.558          & 81.93         & 0.687          & 70.92          & 0.712  & 73.11          & 0.870          & 73.24          & 0.528           & 83.40          & 0.773           & 0.422           \\
RoBERTa$_{BASE}$+AEDA        & 1.753         & 46.14          & 0.559          & 82.03         & 0.686          & 69.96          & 0.734  & 73.12          & 0.577          & 72.66          & 0.410           & 83.29          & 0.780           & 0.435           \\
RoBERTa$_{BASE}$+FreeLB         & 1.738          & 46.77          & 0.575          & \textbf{82.03} & 0.673          & 70.99          & \textbf{0.615}  & 74.48          & 0.542          & 73.83          & 0.495           & 83.59          & 0.793           & 0.501           \\ \hdashline[0.5pt/5pt]
\textbf{Ours}  &&&&&&&&&&&&&&                           \\
RoBERTa$_{BASE}$+RPN            & \textbf{1.717} & \textbf{47.11} & \textbf{0.575} & 81.74          & \textbf{0.650} & \textbf{71.01} & 0.672           & \textbf{74.54} & \textbf{0.528} & \textbf{76.37} & \textbf{0.408}  & \textbf{83.98} & \textbf{0.802}  & \textbf{0.532}  \\ \hline
\end{tabular}
}

\end{table*}
\footnotetext{\url{https://www.kaggle.com/competitions/cola-out-of-domain-open-evaluation}}

Table \ref{Reasult} displays the experimental outcomes, with hyperparameters adjusted according to the approach described in Table \ref{Hyperparameter}. For the EP task, FreeLB employs a learning rate of 1e-4 and a perturbation range of 1e-2, with a random rate of 0.2 in the RPN algorithm and 3 perturbations. The model adopts a learning rate of 3e-5 and a training step epoch of 5 steps. On the ER task, FreeLB uses a learning rate of 1e-3 and a perturbation range of 1e-3, with a random rate of 0.3 in the RPN algorithm and 3 perturbations. The model employs a learning rate of 3e-5 and a training step epoch of 10 steps. On the SA task, FreeLB employs a learning rate of 1e-4 and a perturbation range of 1e-2, with a random rate of 0.3 in the RPN algorithm and 3 perturbations. The model uses a learning rate of 3e-5 and a training step epoch of 3 steps. On the HSD, ID, and OLI tasks, FreeLB adopts a learning rate of 1e-4 and a perturbation range of 1e-2, with a random rate of 0.3 in the RPN algorithm and 3 perturbations. The model uses a learning rate of 3e-5 and a training step epoch of 10 steps. On the CoLA task, FreeLB employs a learning rate of 1e-4 and a perturbation range of 1e-2, with a random rate of 0.3 in the RPN algorithm and 3 perturbations. The model uses a learning rate of 3e-5 and a training step epoch of 10 steps.

\begin{table}[h]
\centering
\caption{\label{Hyperparameter}
For example, hyperparameter search for RPN on the accuracy of validation set of the EP task.
}
\begin{tabular}{|cc|ccc|}
\hline
\multicolumn{2}{|c|}{\multirow{2}{*}{}} & \multicolumn{3}{c|}{$K$}         \\
\multicolumn{2}{|c|}{}                  & 1     & 3              & 5     \\ \hline
\multirow{3}{*}{$\varepsilon$}         & 0.1        & 28.75 & 28.62          & 29.01 \\
                           & 0.2        & 28.45 & \textbf{29.11} & 29.01 \\
                           & 0.5        & 27.45 & 28.10          & 27.11 \\ \hline
\end{tabular}
\end{table}

\begin{figure}[t]
\centering
\includegraphics[scale=0.49]{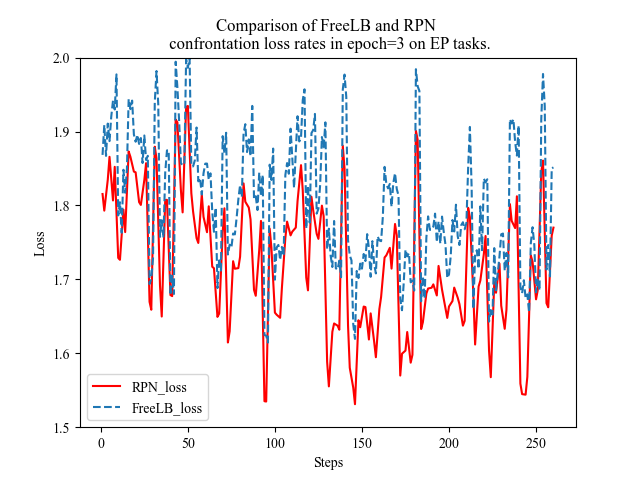}
\caption{Adversarial loss rate of FreeLB and RPN at epoch=3 in the EP dataset.}
\label{adv_loss}
\end{figure}

\paragraph{Results} Table \ref{Reasult} and Figure \ref{adv_loss} shows that the RPN algorithm performs slightly better than other methods in most cases. This may be because the RPN algorithm introduces greater randomness to the samples, making them look like new datasets. However, this randomness may also have a slightly negative impact. If the random drop is too critical, the results may not be consistent with the labels, which would hurt the model's training. On the other hand, this kind of data can bring bad data to the model, preventing such large models from overfitting the training set and making the model more robust.

However, it should be noted that the RPN and FreeLB are not necessarily mutually exclusive, and it is possible to use both methods in combination. This approach will be discussed in a subsequent subsection.

\subsection{FreeLB + RPN}
We combine the RPN algorithm with FreeLB adversarial training, as shown in Section \ref{Application}. 

\begin{table}[h!]
\centering
\caption{\label{RPN+FreeLB}
Results of RPN and FreeLB adversarial training on SA and EP tasks on the test set. Each task was trained 5 times and the mean value was taken. 
}
\resizebox{.95\columnwidth}{!}{
\begin{tabular}{lcccc}
\hline
\multirow{2}{*}{\textbf{Tasks}} & \multicolumn{2}{c}{\textbf{SA}} & \multicolumn{2}{c}{\textbf{EP}} \\ \cline{2-5} 
                                & loss           & acc.            & loss           & acc.            \\ \hline
\textbf{Baseline}              &                 &                &                &                \\
RoBERTa                & 0.650          & 70.51          & 1.755          & 46.11          \\\hdashline[0.5pt/5pt]
\textbf{Other Methods}              &                 &                &                &                \\
RoBERTa+FreeLB         & 0.673          & 70.99          & 1.738          & 46.77          \\ \hdashline[0.5pt/5pt]
\textbf{Ours}              &                 &                &                &                \\
RoBERTa+RPN            & 0.650 & 71.01 & 1.717          & 47.11          \\ 
RoBERTa+FreeLB+RPN     & \textbf{0.649}          & \textbf{71.78}          & \textbf{1.716} & \textbf{47.24} \\ \hline
\end{tabular}
}

\end{table}

In both SA and EP tasks, RPN and FreeLB were performed 3 times, the learning rate of the FreeLB adversarial training and model were the same as set in the previous section, and the RPN random parameter was set to 0.3 ($\varepsilon = 0.3$). 

\paragraph{Results} The results presented in Table \ref{RPN+FreeLB} demonstrate that the combination of FreeLB and RPN can yield improved performance on certain tasks. Specifically, the RPN and FreeLB algorithms used in tandem yielded significantly better results than the original model without the data augmentation algorithm in the EP task. This suggests that data augmentation algorithms may be necessary to optimize model performance when working with limited datasets.

\subsection{Traditional Deep Learning Models}
The RPN algorithm is not limited to transformer-based models such as BERT \cite{devlinetal2019bert}, RoBERTa \cite{zhangetal2022virtual}, and XLNet \cite{10.5555/3454287.3454804}. In this subsection, we perform experiments with a traditional model such as TextCNN to validate the effectiveness of our RPN algorithm.

\begin{table}[]
\centering
\caption{\label{CNN}
FreeLB and RPN algorithms on CNN models on EP and STT-2 tasks. The loss and accuracy on the test set. 
}
\resizebox{.95\columnwidth}{!}{
\begin{tabular}{lcccc}
\hline
\multirow{2}{*}{\textbf{Tasks}} & \multicolumn{2}{c}{\textbf{EP}}    & \multicolumn{2}{c}{\textbf{SST-2}} \\ \cline{2-5} 
                                & loss              & acc.            & loss              & acc.            \\ \hline
\textbf{Baseline}              &                 &                &                &                \\
TextCNN                & 1.897561          & 41.77          & 0.424777          & 82.94          \\ \hdashline[0.5pt/5pt]
\textbf{Other Methods}              &                 &                &                &                \\
TextCNN+FreeLB         & 1.891150          & \textbf{41.95} & 0.420455          & 82.75          \\ \hdashline[0.5pt/5pt]
\textbf{Ours}              &                 &                &                &                \\
TextCNN+RPN            & \textbf{1.889302} & 41.83          & \textbf{0.405808} & \textbf{83.10} \\ \hline
\end{tabular}
}

\end{table}

On the EP and SST-2 tasks, we use three convolution kernels of 10, 20, and 30 lengths to do three layers of convolution, total 4,608,600 parameters, and the learning rate of the model is 1e-4 and training 10 epochs. The FreeLB adversarial training and was the same as set in the previous section, and the RPN random parameter was set to 0.3 ($\varepsilon = 0.3$). 

\paragraph{Results} As demonstrated in Table \ref{CNN}, the RPN algorithm can enhance the performance of traditional deep learning models such as TextCNN. Specifically, the TextCNN model augmented with the RPN algorithm achieved a lower loss on the EP task compared to the original TextCNN model on the test set. Moreover, on the SST-2 task, the TextCNN model with the RPN algorithm yielded better results on the test set in terms of both loss and accuracy compared to the original TextCNN model. These results indicate that our algorithm is not limited to transformer-based models, and can be extended to traditional deep learning models as well.

\section{Efficiency}

\begin{table}[]
\centering
\caption{\label{Efficiency}
We conducted a comparison of the time complexities ($\mathcal{O}$) of various data augmentation techniques, denoting the time complexity in samples processing $+$ model training. Here, $k$ represents the number of times each sample needs to be augmented.
}
\resizebox{.95\columnwidth}{!}{
\begin{tabular}{cccccc}
\hline
\textbf{Methods}         & \textbf{EDA} & \textbf{AEDA} & \textbf{Back-translation} & \textbf{FreeLB} & \textbf{RPN} \\\hline
\textbf{$\mathcal{O}$} & $kn+kn$      & $kn+kn$       & $2kn+kn$                  & $1+kn$          & $1+kn$      
\\ \hline
\end{tabular}}

\end{table}

As indicated in Table \ref{Efficiency}, traditional data augmentation methods rely on the row-by-row processing of samples, leading to significant time costs when dealing with large datasets. In contrast, the proposed RPN algorithm offers greater convenience as it does not require processing the original samples or gradient computation of the word embedding layer from the computational graph.

\section{Conclusion}
In conclusion, there are few approaches to data augmentation for natural language processing, and our contribution is a combination of adversarial training and the analysis of word vector features to propose the RPN algorithm. Our approach differs from previous methods in that it does not require the calculation of gradients to determine the next step of the virtual sample. The RPN algorithm is based on word vectors and demonstrates that random modification of word vectors is a highly effective data augmentation technique, producing good results across various sentence-level tasks. Additionally, the RPN algorithm is applicable to most deep learning models.

\section*{Acknowledgment}
This work gratitude to the Overseas Research and Training Program for Outstanding Young Backbone Talents of Colleges and Universities (NO. GXGWFX2019041), the Industrial Collaborative Innovation Fund of Anhui Polytechnic University and Jiujiang District, China (No. 2021CYXTB9), and the Open Project Foundation of Anhui Provincial Engineering Laboratory on Information Fusion and Control of Intelligent Robot, China (No. IFCIR20200001) for their support and funding towards my research project.

\bibliography{anthology,custom}

% Generated by IEEEtran.bst, version: 1.14 (2015/08/26)
\begin{thebibliography}{10}
\providecommand{\url}[1]{#1}
\csname url@samestyle\endcsname
\providecommand{\newblock}{\relax}
\providecommand{\bibinfo}[2]{#2}
\providecommand{\BIBentrySTDinterwordspacing}{\spaceskip=0pt\relax}
\providecommand{\BIBentryALTinterwordstretchfactor}{4}
\providecommand{\BIBentryALTinterwordspacing}{\spaceskip=\fontdimen2\font plus
\BIBentryALTinterwordstretchfactor\fontdimen3\font minus
  \fontdimen4\font\relax}
\providecommand{\BIBforeignlanguage}[2]{{%
\expandafter\ifx\csname l@#1\endcsname\relax
\typeout{** WARNING: IEEEtran.bst: No hyphenation pattern has been}%
\typeout{** loaded for the language `#1'. Using the pattern for}%
\typeout{** the default language instead.}%
\else
\language=\csname l@#1\endcsname
\fi
#2}}
\providecommand{\BIBdecl}{\relax}
\BIBdecl

\bibitem{10.5555/3495724.3496249}
Q.~Xie, Z.~Dai, E.~Hovy, M.-T. Luong, and Q.~V. Le, ``Unsupervised data
  augmentation for consistency training,'' in \emph{Proceedings of the 34th
  International Conference on Neural Information Processing Systems}, ser.
  NIPS'20.\hskip 1em plus 0.5em minus 0.4em\relax Red Hook, NY, USA: Curran
  Associates Inc., 2020.

\bibitem{PGD}
X.~Zhang, J.~Zhao, and Y.~LeCun, ``Character-level convolutional networks for
  text classification,'' in \emph{Proceedings of the 28th International
  Conference on Neural Information Processing Systems - Volume 1}, ser.
  NIPS'15.\hskip 1em plus 0.5em minus 0.4em\relax Cambridge, MA, USA: MIT
  Press, 2015, p. 649–657.

\bibitem{10.1145/3439726}
\BIBentryALTinterwordspacing
S.~Minaee, N.~Kalchbrenner, E.~Cambria, N.~Nikzad, M.~Chenaghlu, and J.~Gao,
  ``Deep learning--based text classification: A comprehensive review,''
  \emph{ACM Comput. Surv.}, vol.~54, no.~3, apr 2021. [Online]. Available:
  \url{https://doi.org/10.1145/3439726}
\BIBentrySTDinterwordspacing

\bibitem{10.5555/1619499.1619510}
R.~Raina, A.~Y. Ng, and C.~D. Manning, ``Robust textual inference via learning
  and abductive reasoning,'' in \emph{Proceedings of the 20th National
  Conference on Artificial Intelligence - Volume 3}, ser. AAAI'05.\hskip 1em
  plus 0.5em minus 0.4em\relax AAAI Press, 2005, p. 1099–1105.

\bibitem{jiaoetal2020tinybert}
\BIBentryALTinterwordspacing
X.~Jiao, Y.~Yin, L.~Shang, X.~Jiang, X.~Chen, L.~Li, F.~Wang, and Q.~Liu,
  ``{T}iny{BERT}: Distilling {BERT} for natural language understanding,'' in
  \emph{Findings of the Association for Computational Linguistics: EMNLP
  2020}.\hskip 1em plus 0.5em minus 0.4em\relax Online: Association for
  Computational Linguistics, Nov. 2020, pp. 4163--4174. [Online]. Available:
  \url{https://aclanthology.org/2020.findings-emnlp.372}
\BIBentrySTDinterwordspacing

\bibitem{Zhu2020FreeLB:}
\BIBentryALTinterwordspacing
C.~Zhu, Y.~Cheng, Z.~Gan, S.~Sun, T.~Goldstein, and J.~Liu, ``Freelb: Enhanced
  adversarial training for natural language understanding,'' in
  \emph{International Conference on Learning Representations}, 2020. [Online].
  Available: \url{https://openreview.net/forum?id=BygzbyHFvB}
\BIBentrySTDinterwordspacing

\bibitem{wangyang2015thats}
\BIBentryALTinterwordspacing
W.~Y. Wang and D.~Yang, ``That{'}s so annoying!!!: A lexical and frame-semantic
  embedding based data augmentation approach to automatic categorization of
  annoying behaviors using {\#}petpeeve tweets,'' in \emph{Proceedings of the
  2015 Conference on Empirical Methods in Natural Language Processing}.\hskip
  1em plus 0.5em minus 0.4em\relax Lisbon, Portugal: Association for
  Computational Linguistics, Sep. 2015, pp. 2557--2563. [Online]. Available:
  \url{https://aclanthology.org/D15-1306}
\BIBentrySTDinterwordspacing

\bibitem{sennrichetal2016improving}
\BIBentryALTinterwordspacing
R.~Sennrich, B.~Haddow, and A.~Birch, ``Improving neural machine translation
  models with monolingual data,'' in \emph{Proceedings of the 54th Annual
  Meeting of the Association for Computational Linguistics (Volume 1: Long
  Papers)}.\hskip 1em plus 0.5em minus 0.4em\relax Berlin, Germany: Association
  for Computational Linguistics, Aug. 2016, pp. 86--96. [Online]. Available:
  \url{https://aclanthology.org/P16-1009}
\BIBentrySTDinterwordspacing

\bibitem{zhang2020back}
Q.~Zhang, X.~Chen, H.~Li, W.~Lin, and L.~Li, ``Back translation sampling by
  self-training for neural machine translation,'' \emph{IEEE/ACM Transactions
  on Audio, Speech, and Language Processing}, vol.~28, pp. 1697--1710, 2020.

\bibitem{weizou2019eda}
\BIBentryALTinterwordspacing
J.~Wei and K.~Zou, ``{EDA}: Easy data augmentation techniques for boosting
  performance on text classification tasks,'' in \emph{Proceedings of the 2019
  Conference on Empirical Methods in Natural Language Processing and the 9th
  International Joint Conference on Natural Language Processing
  (EMNLP-IJCNLP)}.\hskip 1em plus 0.5em minus 0.4em\relax Hong Kong, China:
  Association for Computational Linguistics, Nov. 2019, pp. 6382--6388.
  [Online]. Available: \url{https://aclanthology.org/D19-1670}
\BIBentrySTDinterwordspacing

\bibitem{karimietal2021aedaeasier}
\BIBentryALTinterwordspacing
A.~Karimi, L.~Rossi, and A.~Prati, ``{AEDA}: An easier data augmentation
  technique for text classification,'' in \emph{Findings of the Association for
  Computational Linguistics: EMNLP 2021}.\hskip 1em plus 0.5em minus
  0.4em\relax Punta Cana, Dominican Republic: Association for Computational
  Linguistics, Nov. 2021, pp. 2748--2754. [Online]. Available:
  \url{https://aclanthology.org/2021.findings-emnlp.234}
\BIBentrySTDinterwordspacing

\bibitem{9156446}
K.~Xu, X.~Yang, B.~Yin, and R.~W. Lau, ``Learning to restore low-light images
  via decomposition-and-enhancement,'' in \emph{2020 IEEE/CVF Conference on
  Computer Vision and Pattern Recognition (CVPR)}, 2020, pp. 2278--2287.

\bibitem{9156559}
W.~Yang, S.~Wang, Y.~Fang, Y.~Wang, and J.~Liu, ``From fidelity to perceptual
  quality: A semi-supervised approach for low-light image enhancement,'' in
  \emph{2020 IEEE/CVF Conference on Computer Vision and Pattern Recognition
  (CVPR)}, 2020, pp. 3060--3069.

\bibitem{perez2014automatic}
V.~Perez-Rosas, C.~Banea, and R.~Mihalcea, ``Automatic generation of rephrased
  texts for paraphrase detection,'' in \emph{Proceedings of the 14th Conference
  of the European Chapter of the Association for Computational Linguistics},
  2014, pp. 683--692.

\bibitem{karpukhin2021denoising}
V.~Karpukhin, M.~Wortsman, K.~Tu, X.~Li, K.~Gimpel, D.~He, and K.~Cho,
  ``Denoising diffusion probabilistic models for text generation,'' in
  \emph{Conference on Empirical Methods in Natural Language Processing}, 2021,
  pp. 2524--2534.

\bibitem{madry2018towards}
\BIBentryALTinterwordspacing
A.~Madry, A.~Makelov, L.~Schmidt, D.~Tsipras, and A.~Vladu, ``Towards deep
  learning models resistant to adversarial attacks,'' in \emph{International
  Conference on Learning Representations}, 2018. [Online]. Available:
  \url{https://openreview.net/forum?id=rJzIBfZAb}
\BIBentrySTDinterwordspacing

\bibitem{DBLP:journals/corr/GoodfellowSS14}
\BIBentryALTinterwordspacing
I.~J. Goodfellow, J.~Shlens, and C.~Szegedy, ``Explaining and harnessing
  adversarial examples,'' in \emph{3rd International Conference on Learning
  Representations, {ICLR} 2015, San Diego, CA, USA, May 7-9, 2015, Conference
  Track Proceedings}, Y.~Bengio and Y.~LeCun, Eds., 2015. [Online]. Available:
  \url{http://arxiv.org/abs/1412.6572}
\BIBentrySTDinterwordspacing

\bibitem{wang-etal-2018-glue}
\BIBentryALTinterwordspacing
A.~Wang, A.~Singh, J.~Michael, F.~Hill, O.~Levy, and S.~Bowman, ``{GLUE}: A
  multi-task benchmark and analysis platform for natural language
  understanding,'' in \emph{Proceedings of the 2018 {EMNLP} Workshop
  {B}lackbox{NLP}: Analyzing and Interpreting Neural Networks for {NLP}}.\hskip
  1em plus 0.5em minus 0.4em\relax Brussels, Belgium: Association for
  Computational Linguistics, Nov. 2018, pp. 353--355. [Online]. Available:
  \url{https://aclanthology.org/W18-5446}
\BIBentrySTDinterwordspacing

\bibitem{barbieri-etal-2020-tweeteval}
\BIBentryALTinterwordspacing
F.~Barbieri, J.~Camacho-Collados, L.~Espinosa~Anke, and L.~Neves,
  ``{T}weet{E}val: Unified benchmark and comparative evaluation for tweet
  classification,'' in \emph{Findings of the Association for Computational
  Linguistics: EMNLP 2020}.\hskip 1em plus 0.5em minus 0.4em\relax Online:
  Association for Computational Linguistics, Nov. 2020, pp. 1644--1650.
  [Online]. Available: \url{https://aclanthology.org/2020.findings-emnlp.148}
\BIBentrySTDinterwordspacing

\bibitem{warstadtetal2019neural}
\BIBentryALTinterwordspacing
A.~Warstadt, A.~Singh, and S.~R. Bowman, ``Neural network acceptability
  judgments,'' \emph{Transactions of the Association for Computational
  Linguistics}, vol.~7, pp. 625--641, 2019. [Online]. Available:
  \url{https://aclanthology.org/Q19-1040}
\BIBentrySTDinterwordspacing

\bibitem{socheretal2013recursive}
\BIBentryALTinterwordspacing
R.~Socher, A.~Perelygin, J.~Wu, J.~Chuang, C.~D. Manning, A.~Ng, and C.~Potts,
  ``Recursive deep models for semantic compositionality over a sentiment
  treebank,'' in \emph{Proceedings of the 2013 Conference on Empirical Methods
  in Natural Language Processing}.\hskip 1em plus 0.5em minus 0.4em\relax
  Seattle, Washington, USA: Association for Computational Linguistics, Oct.
  2013, pp. 1631--1642. [Online]. Available:
  \url{https://www.aclweb.org/anthology/D13-1170}
\BIBentrySTDinterwordspacing

\bibitem{devlinetal2019bert}
\BIBentryALTinterwordspacing
J.~Devlin, M.-W. Chang, K.~Lee, and K.~Toutanova, ``{BERT}: Pre-training of
  deep bidirectional transformers for language understanding,'' in
  \emph{Proceedings of the 2019 Conference of the North {A}merican Chapter of
  the Association for Computational Linguistics: Human Language Technologies,
  Volume 1 (Long and Short Papers)}.\hskip 1em plus 0.5em minus 0.4em\relax
  Minneapolis, Minnesota: Association for Computational Linguistics, Jun. 2019,
  pp. 4171--4186. [Online]. Available: \url{https://aclanthology.org/N19-1423}
\BIBentrySTDinterwordspacing

\bibitem{zhangetal2022virtual}
\BIBentryALTinterwordspacing
D.~Zhang, W.~Xiao, H.~Zhu, X.~Ma, and A.~Arnold, ``Virtual augmentation
  supported contrastive learning of sentence representations,'' in
  \emph{Findings of the Association for Computational Linguistics: ACL
  2022}.\hskip 1em plus 0.5em minus 0.4em\relax Dublin, Ireland: Association
  for Computational Linguistics, May 2022, pp. 864--876. [Online]. Available:
  \url{https://aclanthology.org/2022.findings-acl.70}
\BIBentrySTDinterwordspacing

\bibitem{10.5555/3454287.3454804}
Z.~Yang, Z.~Dai, Y.~Yang, J.~Carbonell, R.~Salakhutdinov, and Q.~V. Le,
  \emph{XLNet: Generalized Autoregressive Pretraining for Language
  Understanding}.\hskip 1em plus 0.5em minus 0.4em\relax Red Hook, NY, USA:
  Curran Associates Inc., 2019.

\end{thebibliography}

\end{document}